\documentclass[10pt,twocolumn,letterpaper]{article}

\usepackage[pagenumbers]{cvpr} 

\usepackage{graphicx}
\usepackage{amsmath}
\usepackage{amssymb}
\usepackage{bm}
\usepackage{booktabs}
\DeclareMathOperator*{\argmax}{arg\,max}

\usepackage[pagebackref,breaklinks,colorlinks]{hyperref}

\usepackage[capitalize]{cleveref}
\crefname{section}{Sec.}{Secs.}
\Crefname{section}{Section}{Sections}
\Crefname{table}{Table}{Tables}
\crefname{table}{Tab.}{Tabs.}

\def\cvprPaperID{*****} 
\def\confName{CVPR}
\def\confYear{2023}

\begin{document}

\title{Generative Robust Classification}

\author{Xuwang Yin\\
University of Virginia\\
{\tt\small xy4cm@virginia.edu}
}
\maketitle

\begin{abstract}
Training adversarially robust discriminative (i.e., softmax) classifier has been the dominant approach to robust classification. Building on recent work on adversarial training (AT)-based generative models, we investigate using AT to learn unnormalized class-conditional density models and then performing generative robust classification. Our result shows that, under the condition of similar  model capacities, the generative robust classifier achieves comparable performance to a baseline softmax robust classifier when the test data is clean or when the test perturbation is of limited size, and much better performance when the test perturbation size exceeds the training perturbation size. The generative classifier is also able to generate samples or counterfactuals that more closely resemble the training data, suggesting that the generative classifier can better capture the class-conditional distributions. In contrast to standard discriminative adversarial training where advanced data augmentation techniques are only effective when combined with weight averaging, we find it straightforward to apply advanced data augmentation to achieve better robustness in our approach. Our result suggests that the generative classifier is a competitive alternative to robust classification, especially for problems with limited number of classes.

\end{abstract}

\section{Introduction}
\label{sec:intro}
\textit{Discriminative classification} and \textit{generative classification} are two different approaches to solving classification problems. In discriminative classification, we directly model the posterior distribution of the class labels $p({k}|\bm{x})$. This posterior distribution is typically obtained by applying the \textit{softmax} function to the logit outputs of the classifier. 
\begin{equation}
	p({k}|{\bm{x}}) = \frac{\exp (a_k)}{\sum_{j=1}^{K} \exp (a_j)},
\end{equation}
where $a_k$ is the model's $k$-th output. 
Because the discriminative classifier makes use of the softmax function, it is also known as\textit{ softmax classifier}. The discriminative classifier is in fact modeling the decision boundary between different classes. 
An alternative approach is to first model the class-conditional distributions $p(\bm{x}|{k})$ and then obtain the posterior distribution using Bayes' theorem
\begin{equation}
	p({k}|{\bm{x}}) = \frac{p({\bm{x}}|{k})p({k})}{p({\bm{x}})},
\end{equation}
where the prior distribution $p({k})$  can be estimated from the fractions of the training set data points in each of the classes. This is the generative approach to solving classification problems. It seems that in order to solve the classification problem, the generative classifier needs to solve a more complex problem of modeling the class-conditional distributions. Indeed, the class-conditional distribution may contain a lot of structure that has little effect on $p({k}|{\bm{x}})$. But directly modeling   $p(\bm{x}|{k})$ can be advantageous when we want to use  $p(\bm{x}|{k})$ to detect out-of-distribution inputs.

Deep neural network-based discriminative classifier is the dominant approach to solving classification problems. A widely known issue of this approach is the existence of adversarial examples. 
Various defense mechanisms have been proposed to address adversarial examples, with \textit{Adversarial training (AT)} being the most successful one. AT improves the classifier's robustness by training the classifier against adversarially perturbed inputs. One issue with AT is that adversarially robust classifiers tend to have lower accuracy on clean data than standard, non-robust classifiers. \cite{tsipras2018robustness} conjecture that the reduced accuracy on clean data is the consequence of robust classifiers learning fundamentally different features than standard classifiers. 
Recent work on AT has been focus on using data augmentation (combined with model weight averaging~\cite{izmailov2018averaging})~\cite{rebuffi2021fixing}, synthetic training data produced by generative models~\cite{rebuffi2021fixing,gowal2021improving}, and high capacity models~\cite{alayrac2019labels, xie2019intriguing, gowal2020uncovering} to improve the standard and robust accuracies.
Discriminative classifier also has the issue of producing overconfident predictions on out-of-distribution inputs.
\cite{augustin2020adversarial} propose to address this issue by combining in- and out-distribution adversarial training to enforce low confidence on adversarial out-of-distribution samples.

Compared to discriminative robust classification, generative robust classification is a less-explored area.
Robust generative classification  requires the density model $p_{\bm{\theta}}({\bm{x}}|{k})$ to have low likelihood outputs on regular as well as adversarial out-of-distribution samples, which is not achievable with existing density models. 
In this work we explore modeling $p({\bm{x}}|{k})$ with the AT generative model proposed in \cite{yin2022learning}, and then using these conditional models to perform robust generative classification. We demonstrate that this robust generative classifier achieves comparable standard and robust classification accuracies to state-of-the-art softmax robust classifiers, and at the same time is more interpretable. 
One issue with the proposed generative classifier is that it does not easily scale to problems with many classes. However, we note there are many practical problems with limited number of classes~\cite{Dua2019}, and our approach may find applications in this kind of problems.

We note that generative (robust) classification has also been investigated in \cite{Yin2020GAT}. In terms of the formulation of the generative classifier,  both \cite{Yin2020GAT} and our work use AT to train a binary classifier to model the unnormalized density function of the class-conditional data. The key difference is that we do not assume that the normalizing constants of the density functions of different classes to be equal, and instead treat the normalizing constants as learnable parameters of the generative classifier. In terms of implementation, \cite{Yin2020GAT}'s generative classifier is ten times larger than the softmax robust classifier in terms of capacity~\cite{madry2017towards}, and yet its performance is much worse~\cite{tramer2020adaptive}.
In summary, our specific contributions are:
\begin{itemize}
	\item We show that when the overall model capacities are similar, the generative classifier achieve comparable standard and robust accuracies to a baseline softmax robust classifier. In particular, the generative classifier is much more robust when the test perturbation is large.
	\item We study the interpretability of the generative classifier and show that the generative classifier is more interpretable in terms of the qualities of generated samples and counterfactuals.
	\item We propose to combine in- and out-distribution adversarial training to improve the model robustness.
	\item We perform comprehensive ablation to study how calibration, model capacity, weight decay regularization, training perturbation size, data augmentation, and combining in- and out-distribution AT affect the model performance. 
\end{itemize}


\section{Generative Robust Classification}
In a $K$ class classification problem, the proposed generative classifier consists of $K$ binary classifiers, with the $k$-th binary classifier  trained to distinguish clean data of class $k$ from adversarially perturbed out-of-distribution data.
Following~\cite{yin2022learning}, denote the data distribution of class $k$ as $p_{k}$, the out-of-distribution dataset as $p_0$, then the $k$-th binary classifier $D_k:\mathcal{X} \subseteq \mathbb{R}^d \rightarrow [0,1]$  is trained by maximizing the following objective:
\begin{equation}
	\label{eq:gat-obj-d}
	\begin{split}
	J({D_k}) =  ~&\mathbb{E}_{{\bm{x}} \sim p_{k}}[\log {D_k}({\bm{x}})]  \\ & + \mathbb{E}_{{\bm{x}} \sim p_{0}}[\min_{\bm{x}'\in\mathbb{B}(\bm{x},\epsilon)}\log (1 - {D_k}(\bm{x}')))],
	\end{split}
\end{equation}
where $\mathbb{B}(\bm{x},\epsilon)$ is a neighborhood of $\bm{x}$: $\mathbb{B}(\bm{x},\epsilon)=\{\bm{x}' \in \mathcal{X}: \| \bm{x}'-\bm{x}\|_2 \leq \epsilon \}$. 
$D_k$ is defined as $D_k(\bm{x}) = \sigma(d_k(\bm{x}))$, where $\sigma$ is the logistic sigmoid function, and $d_k:\mathbb{R}^d\rightarrow \mathbb{R}$ is a neural network with a single output node.

As is discussed in \cite{yin2022learning}, training with a larger perturbation $\epsilon$ causes the model to learn to generate more realistic and diverse samples of $p_k$.
Unfortunately, there is typically a trade-off between $D_k$'s discriminative capability and its generative capability. In this work we focus on optimizing the model's (standard and robust) discriminative performance by choosing an approximate $\epsilon$ to train the model. \cite{yin2022learning} also suggests that the $p_0$ should be diverse in order for $D_k$ to better capture $p_k$. However, in order to have a fair comparison with the discriminative classifier which does not use additional data, we use the the mixture distribution of other classes as the out-of-distribution dataset: $p_0=p_{\backslash k} =\frac{1}{K-1}\sum_{i=1,...,K, i\neq k} p_i$.


Following \cite{yin2022learning}, we interpret $d_k(\bm{x})$ as an unnormalized density model of $p(\bm{x}|k)$.
We can obtain the normalized density function by
\begin{equation}p(\bm{x}| k) = \frac{\exp(d_k(\bm{x}))}{Z_k},
\end{equation} with $Z_k$ being the partition function: 
\begin{equation}
	Z_k = \int\exp(d_k(\bm{x}))d\bm{x}.
\end{equation}
We can then apply the Bayes classification rule to obtain a \textit{generative classifier} $g(\bm{x}): \mathbb{R}^d \to \{1, ..., K\}$: 
\begin{equation}
	\label{eq:generative_classifier}
	\begin{split}
		g(\bm{x}) & = \argmax_k p(k|\bm{x}) \\
		& = \argmax_k \frac{p(\bm{x}|k)p(k)}{p(\bm{x})}  \\ 
		& = \argmax_k \frac{\exp (d_k(\bm{x}))p(k)}{ Z_k},
	\end{split}
\end{equation}
\subsection{Calibration}
\label{sec:calibration}
In general, $Z_k$ is  intractable, and $Z_k$ of different classes are not equal, so $g(\bm{x})$ is also  intractable.
While $Z_k$ cannot be computed directly, we can treat $Z_k$ as learnable parameters of the generative classifier, and then obtain their values by optimizing the generative classifier's performance on a validation set.
Absorbing $Z_k$ into the $ \exp $ and assuming that $p(k)$ of different classes are equal, then the generative classifier can be simplified as 
\begin{equation}
	\label{eq:generative_classifier_logit}
	g(\bm{x})= \argmax_k  d_k(\bm{x}) + c_k,
\end{equation}
where the calibration constants $\{c_1,...,c_K\}$ can be learned on a validation set.


\subsection{Combining in- and out-distribution adversarial training} 
The robustness of the generative classifier can be further improved by combining in- and out-distribution adversarial training. 
Consider an adversarially perturbed sample $ \bm{x}'=\bm{x}+\bm{\delta} $, where $\bm{x}$ is a clean sample of class $k$, and  $ \bm{\delta} $ is an adversarial perturbation intended to cause misclassification. 
In order  to correctly classify $ \bm{x}' $, the generative classifier \cref{eq:generative_classifier_logit} needs satisfy
\begin{equation}\label{eq:inadv-req}
	\forall i \in \{1, ..., K\} \backslash \{k\} \quad  d_k(\bm{x}+\bm{\delta}) +c_k > d_i(\bm{x}+\bm{\delta})+c_i.
\end{equation}
However, in \cref{eq:gat-obj-d}, $d_k$ is only trained to have high outputs on clean samples of $p_k$, not perturbed samples of $p_k$. We can increase $d_k$'s outputs on perturbed samples of $p_k$ by also performing in-distribution adversarial training:
\begin{equation}
	\label{eq:gat-obj-inclass}
	\begin{split}
	J({D_k}) =~ &\mathbb{E}_{\mathrm{\bm{x}} \sim p_{k}}[ \min_{\bm{x}'\in\mathbb{B}(\bm{x},\epsilon_1)}\log {D_k}({\bm{x}'})] \\& +  \mathbb{E}_{\mathrm{\bm{x}} \sim p_{\backslash k}}[\min_{\bm{x}'\in\mathbb{B}(\bm{x},\epsilon_2)}\log (1 - {D_k}(\bm{x}')))].
\end{split}
\end{equation} 
By explicitly training $d_k$ to have high outputs on  \textit{adversarially perturbed samples} of class $k$, we make  \cref{eq:inadv-req} easier to satisfy. 
In \cref{sec:generative_classification_ablation} we provide an ablation on combining in- and out-distribution adversarial training.
	
\section{Experiments}
\subsection{Training setup}
We evaluate the generative classifier on the CIFAR-10 dataset (ten classes problem), a widely used benchmark for robust classification~\cite{croce2020robustbench}. 
The proposed generative classifier consists of $K$ binary classifiers in a $K$ class classification problem. In order to have fair comparison with the discriminative approach which typically uses a single model, we limit the capacity of the binary classifier so that the overall capacities of these two approaches are similar. 
Specifically, we use a customized model ``ResNet18Thinner'' (see also in \cref{fig:capacity_ablation}) to train the binary classifiers. The ResNet18Thinner model is an ResNet18 architecture with a width multiplier of 0.5, and has 10M parameters. The baseline discriminative robust classifier~\cite{robustness} is based on the ResNet50 architecture which has  90MB parameters.

All the binary classifier are trained with the SGD optimizer using a batch size of 128 and weight decay $10^{-4}$ for 2000 epochs. For classes 2, 3, 4, 5, 7, 8, 9 we use a starting learning rate of 0.1, and for classes 0, 1, 6 we find the 0.1 learning rate being too high and instead use a starting learning rate of 0.05. The learning rate is reduced to 0.01 after epoch 1500. The training perturbation size is 0.3 ($L_2$ norm), in contrast to the perturbation size of 0.5 used in training the softmax robust classifier. Following the common practice in the adversarial machine learning literature~\cite{rice2020overfitting}, we use early stopping on the test set to select the model.
\subsection{Evaluation}
We evaluate the generative classifier and softmax robust classifier on the clean test set (clean accuracy) and the adversarially perturbed test set (robust accuracy). The adversarial perturbations are computed by performing untargeted adversarial attack against the classifiers. Given a test sample $\bm{x}\in \mathbb{R}^d$ and its label $y$, for the softmax robust classifier $f$, the adversarial perturbation $\bm{\delta}^*$ is computed by solving 
\begin{equation}
	\bm{\delta}^* = \argmax_{\bm{\delta}\in \mathbb{R}^d, \|\bm{\delta}\|_p \le\epsilon } L (f(\bm{x}+\bm{\delta}), y),
\end{equation}
where $L$ is the cross-entropy loss. We solve this optimization using the PGD attack~\cite{madry2017towards}. For the generative classifier, it has been shown that directly attacking the cross-entropy loss is suboptimal~\cite{tramer2020adaptive}.
We instead use the attack proposed by \cite{tramer2020adaptive}. The attack first compute  
\begin{equation}
	\bm{\delta}_k^* = \argmax_{\bm{\delta}\in \mathbb{R}^d, \|\bm{\delta}\|_p \le\epsilon }d_k(\bm{x}+\bm{\delta})
\end{equation}
for $k = 1, ..., K, k\neq y$ and then compute $\bm{\delta}^*$ by
\begin{equation}
		\bm{\delta}^* = \argmax_{k = 1, ..., K, k\neq y} d_k(\bm{x}+\bm{\delta}_k^*).
\end{equation}

\subsection{Results}

\subsubsection{Standard accuracy and robust accuracy}
\begin{figure}[h!]
	\centering
	\begin{subfigure}[t]{0.49\linewidth}
		\centering
		\includegraphics[width=\linewidth]{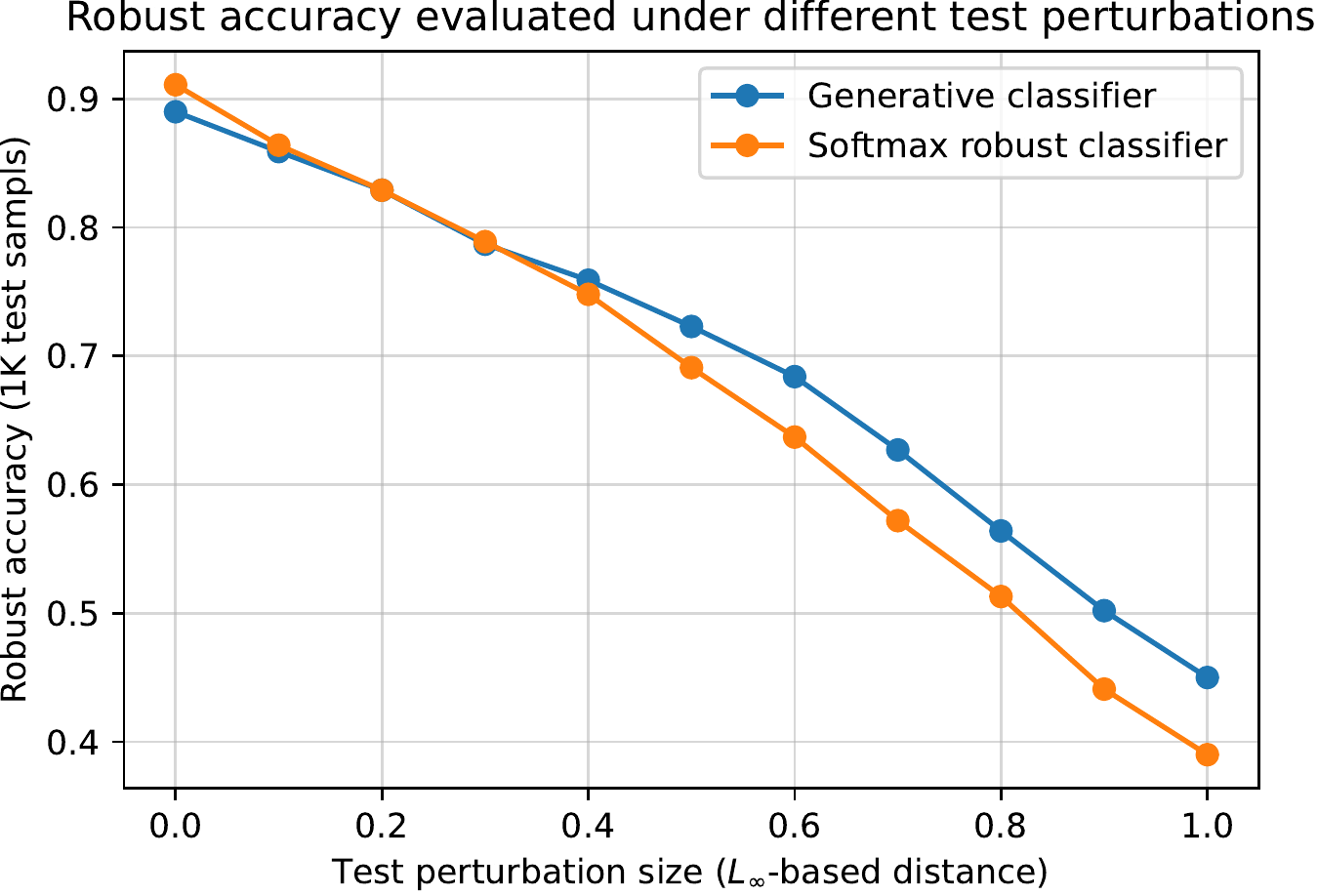}	
	\end{subfigure}
	\begin{subfigure}[t]{0.49\linewidth}
		\centering
		\includegraphics[width=\linewidth]{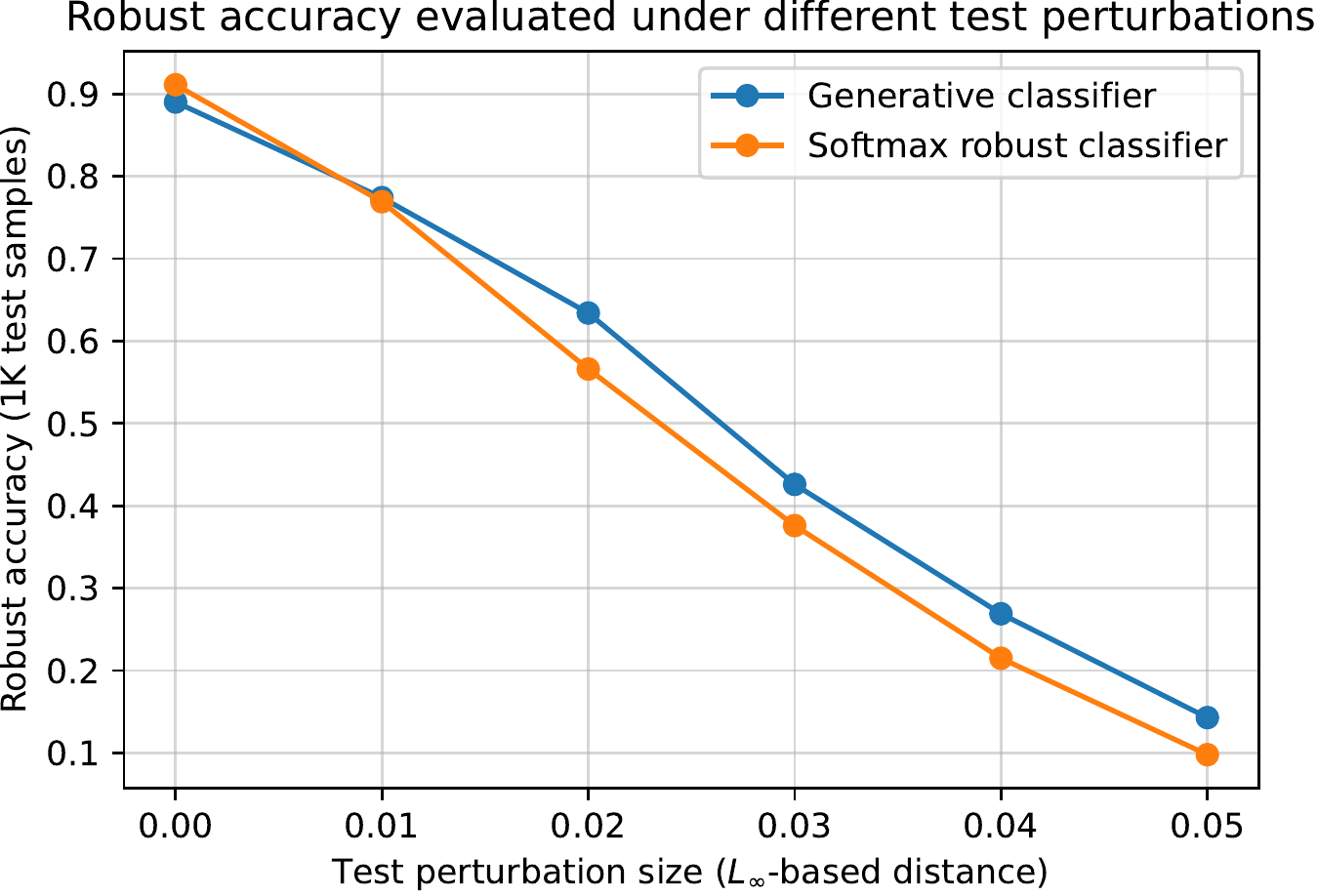}	
	\end{subfigure}
	
	\caption{The performance of the generative classifier and the softmax robust classifier tested under different test perturbation sizes.}
	\label{fig:cifar10_acc}
\end{figure}

\cref{tab:cifar10_acc} shows that the generative classifier has lower standard accuracy but higher robust accuracy compared to the softmax robust classifier. 
To further investigate this phenomenon, we evaluate these two classifiers using the test set perturbed with different levels of adversarial noise. \cref{fig:cifar10_acc}  shows that when there is no perturbation or when the test perturbation is small, the softmax robust classifier has better performance, and when the test perturbation is of moderate size or large size, the generative classifier outperforms the softmax robust classifier, and the larger the test perturbation, the larger the performance gap. 

In adversarial machine learning, it has been observed that there is a trade-off between standard accuracy and robust accuracy~\cite{tsipras2018robustness} --- training with adversarially perturbed data allows the classifier to learn more robust and  semantically meaningful  features to achieve improved robust accuracy at the expense of the standard accuracy. ~\cite{robustness} further shows that larger training perturbations causes the model to perform better under large test perturbations but worse under small or no test perturbations. 
As~\cref{fig:perturbation_ablation} suggests, this trade-off between standard performance and robust performance also exists when we use \cref{eq:gat-obj-d} to train the binary classifiers.
Note that the training perturbation of the generative classifier is already much smaller than that of the softmax robust classifier (0.3 vs. 0.5).
Based on these results, we conjecture that the generative classifier can learn more  semantically meaningful features by capturing the class-conditional distributions and therefore has a better robust accuracy. Indeed, \cref{tab:fid_cifar10_gen}, \cref{fig:global_inte}, and \cref{fig:inter_counterfactuals} show that the generated samples of the generative classifier have better quality than that of the softmax robust classifier, suggesting that the generative classifier has better generative properties.
\begin{table}[h!]
	\centering
	\resizebox{\linewidth}{!}{%
			\begin{tabular}{@{}lcc@{}}
				\toprule
				Model                                                               & Standard accuracy & Robust accuracy  \\ \midrule
				Generative classifier                                               & 88.12\%           & 72.27\%         \\
				Softmax robust classifier\cite{robustness}                                           & 90.83\%           & 70.17\%         \\ \bottomrule
			\end{tabular}%
		}
		\caption{CIFAR10 standard accuracy and robust accuracy ($\epsilon_\textrm{test}=0.5$, $L_2$ norm). }
		\label{tab:cifar10_acc} 
\end{table}

\subsubsection{Interpretability}
We also study the interpretability the generative classifier. We use two approaches to evaluate the classifier's interpretability. First, we visualize the ``learned concepts'' of the classifier by generating synthetic samples that cause high-confidence predictions. This allows us to understand how the model make prediction in general and can be considered as a \textit{global interpretability} method~\cite{molnar2020interpretable}. 
Second, we consider \textit{counterfactual explanation}~\cite{wachter2017counterfactual} which is also used by previous work~\cite{augustin2020adversarial} to evaluate the interpretability of adversarially robust classifiers. 
To access the quality of generated samples (or counterfactuals), we compute the FID~\cite{heusel2017gans} between the training samples and  the generated samples. FID is a distance metric that measures the similarity between two sets of data samples, and is widely used to evaluate the quality of images produced by generative models.

\begin{figure}[h!]
	\centering
	\begin{subfigure}[t]{0.32\linewidth}
		\centering
		\includegraphics[width=\linewidth]{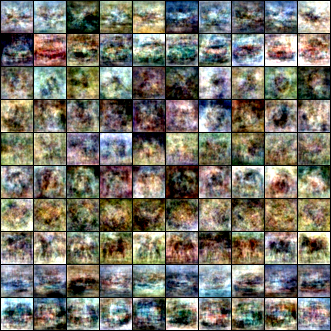}	
		\caption{Seed images}
	\end{subfigure}
	\begin{subfigure}[t]{0.32\linewidth}
		\centering
		\includegraphics[width=\linewidth]{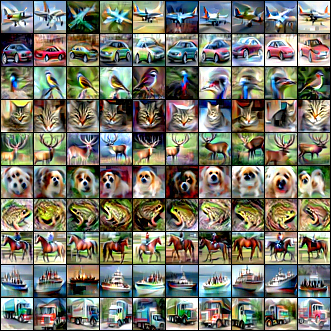}	
		\caption{Generated samples (generative classifier)}
	\end{subfigure}
	\begin{subfigure}[t]{0.32\linewidth}
		\centering
		\includegraphics[width=\linewidth]{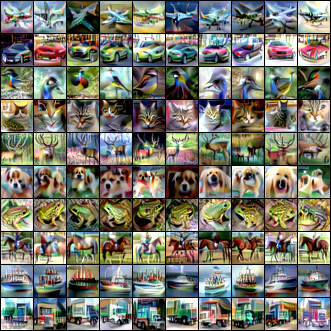}	
		\caption{Generated samples (softmax  classifier)}
	\end{subfigure}
	
	\caption{Seed images and generated images generated by performing targeted adversarial attack on the generative classifier and the robust discriminative classifier. The PGD attack is $L_2$-based attack of steps 10 and step-size 1.0.}
	\label{fig:global_inte}
\end{figure}

\noindent\textbf{Generated samples.} To generate samples that cause high-confidence predictions,   we follow~\cite{santurkar2019image} and fit a multivariate normal distribution to the class-conditional data and then generate seed images  by sampling from the this normal distribution. We then generate samples by performing targeted PGD attacks against the model with the seed images. We use a large perturbation limit so that the generated samples cause high-confidence predictions.
\cref{fig:global_inte} shows the seed images and generated images of the generative classifier and the robust discriminative classifier. 
In both cases, the generated samples resemble the class-conditional data, suggesting the classifiers have captured the high-level features of the target classes. Qualitatively, generated  samples of the generative classifier have less artifacts and the foreground objects are more recognizable.
To provide a quantitative assessment, we compute the FID of the generated samples. \cref{tab:fid_cifar10_gen} shows that the generated samples of the generative classifier has lower FID than that of the softmax robust classifier under different configurations of the PGD attack, suggesting that the generated samples of the generative classifier more closely resemble the training data.
We note that samples generated with different  PGD attacks tend to have different FIDs, but samples produced by the generative classifier have consistently lower FIDs.

\begin{figure}[h!]
	\centering
	\begin{subfigure}[t]{0.32\linewidth}
		\centering
		\includegraphics[width=\linewidth]{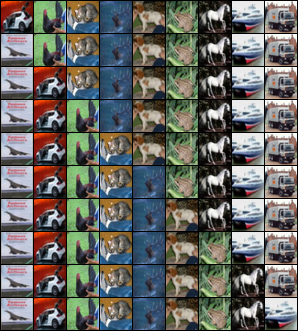}	
		\caption{Seed images}
	\end{subfigure}
	\begin{subfigure}[t]{0.32\linewidth}
		\centering
		\includegraphics[width=\linewidth]{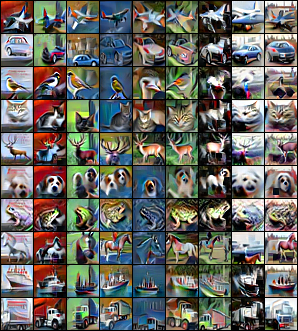}	
		\caption{Counterfactuals for the generative classifier}
	\end{subfigure}
	\begin{subfigure}[t]{0.32\linewidth}
		\centering
		\includegraphics[width=\linewidth]{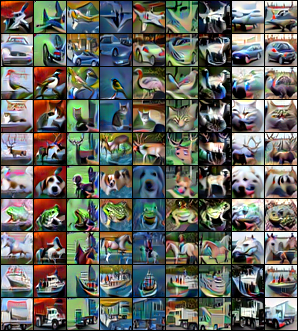}	
		\caption{Counterfactuals for the softmax robust classifier}
	\end{subfigure}
	
	\caption{Seed images and counterfactuals generated by performing targeted adversarial attack on the generative classifier and the robust discriminative classifier. The PGD attack is $L_2$-based attack of steps 10 and step-size 1.0.}
	\label{fig:inter_counterfactuals}
\end{figure}

\noindent\textbf{Counterfactuals.} The idea of counterfactual explanation is to find the minimum change to an input sample such that the modified sample is classified to a predefined target class. To generate counterfactuals we take input samples and then perform targeted adversarial attacks against the classifier. For standard, non-robust models, this process would lead to adversarial examples~\cite{goodfellow2014explaining} which do not have visually meaningful changes. If on the other hand the classifier has captured the class-specific features, then the classifier's decisions change only when class-specific features appear in the attacked data.
\cref{fig:inter_counterfactuals} shows both the generative classifier and softmax robust classifier are able to generate plausible counterfactual explanations. The counterfactuals generated with the two approaches are qualitative similar in terms of the emergence of class specific  features. We again use FID as a quantitative measure and \cref{tab:fid_cifar10_gen} shows that the counterfactuals of the generative classifier has better quality in terms of the FID score.

\begin{table}[h!]
	\begin{center}
		\resizebox{\linewidth}{!}{%
\begin{tabular}{@{}lcc@{}}
	\toprule
	& Generated samples & Counterfactuals \\ \midrule
	\multicolumn{3}{l}{$L_2$ PGD attack of steps 7 and step-size 1.0}                 \\ \midrule
	Generative classifier                       & 50.38             & 37.92           \\
	Softmax robust classifier~\cite{robustness} & 61.84             & 43.65           \\ \midrule
	\multicolumn{3}{l}{$L_2$ PGD attack of steps 10 and step-size 1.0}                \\ \midrule
	Generative classifier                       & 54.78             & 42.33           \\
	Softmax robust classifier~\cite{robustness} & 66.15             & 50.85           \\ \bottomrule
\end{tabular}%
		}
	\end{center}
		\caption{FID of generated samples  and counterfactuals of the generative classifier and softmax robust classifier under different configurations of the PGD attack}
\label{tab:fid_cifar10_gen}  
\end{table}

\subsection{Ablation study}
\label{sec:generative_classification_ablation}
In this section we study how calibration, model capacity, weight decay regularization, training perturbation size, data augmentation, and combining in- and out-distribution adversarial training affect the model performance. 
To make the analysis more concrete, we study how the above factors affect the performance of individual binary classifiers.  We consider both the normal and adversarial scenarios, where in the normal scenario we evaluate the binary classifier on the task of separating clean in-distribution test samples from out-distribution test samples, and in the adversarial scenario we evaluate the binary classifier on the task of separating adversarially perturbed in-distribution and out-distribution test samples. We use the AUC (area under the ROC curve) score as the performance metric.

\subsubsection{Calibration}
\begin{figure}[h!]
	\centering
	\resizebox{0.8\linewidth}{!}{%
		\centering
		\includegraphics[width=\textwidth]{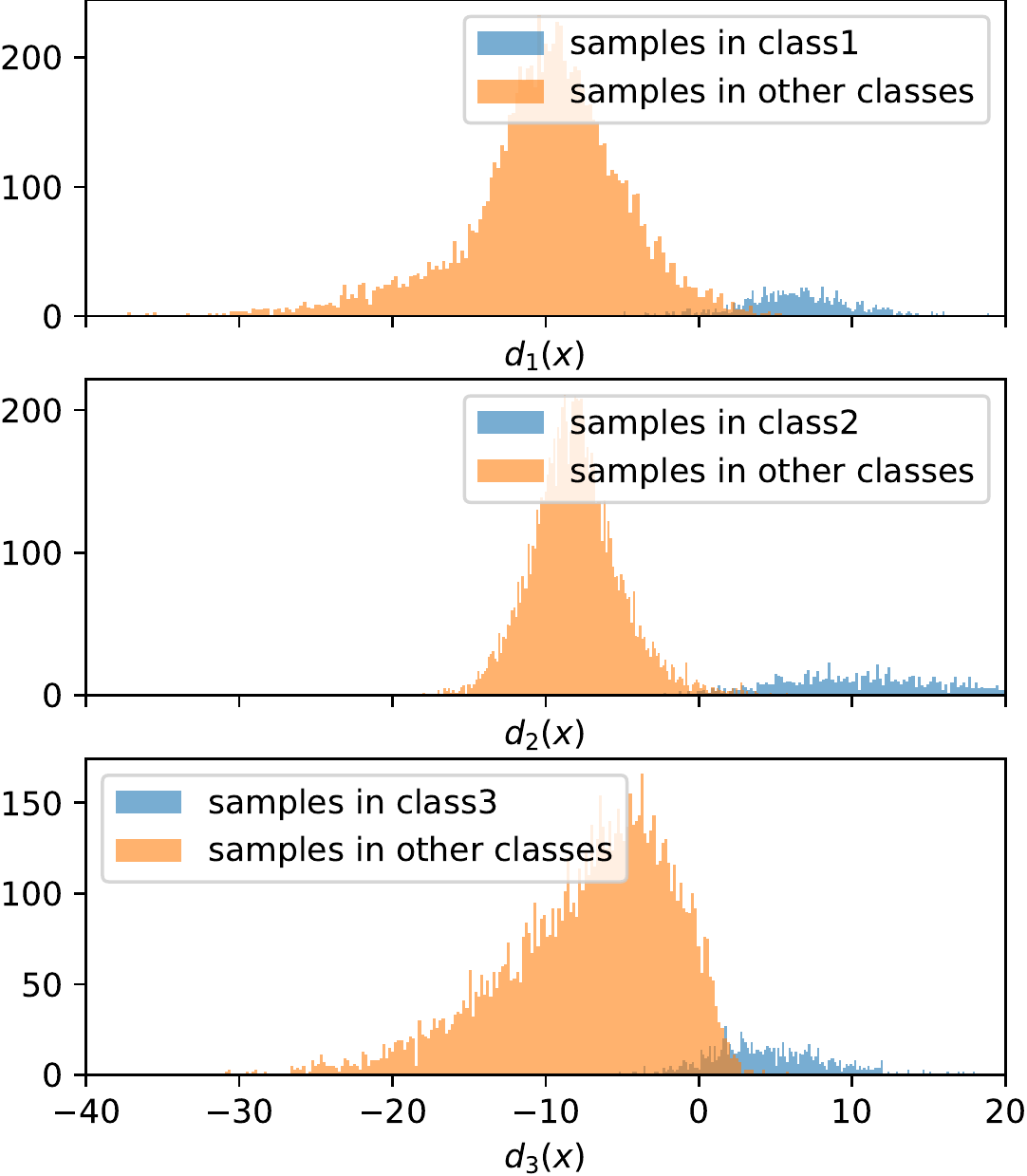}
	}
	\caption{Histogram of $ d_k $'s outputs on samples of class $ k $ and samples of other classes, $ k=1,2,3 $. (Note that the x axes of the subplots are shared.)} 
	\label{fig:logit_hist}
\end{figure}

The calibration parameters $\{c_1,...,c_K\}$ are learned by minimizing the cross-entropy loss of the generative classifier \cref{eq:generative_classifier_logit} on the training set of CIFAR10. 
We use early stopping on the test set to select $\{c_1,...,c_K\}$.
The binary classifiers turn out to be well-calibrated, and  the calibration does not have much effect on the generative classifier's performance (\cref{tab:calibration}). \cref{tab:calibrated_vals} shows the values of the learned  $\{c_1,...,c_K\}$ (the initial values of $\{c_1,...,c_K\}$ are set to zeros).
\cref{fig:logit_hist} shows the histogram of $ d_k $'s outputs on samples of class $ k $ and samples of other classes. It can be seen that the learned  $\{c_1,...,c_K\}$ are tiny compared to the outputs of $ d_k $, so they do not have much effect on \cref{eq:generative_classifier_logit}.
We note that although the calibration does not have much effect on the evaluated dataset (CIFAR-10), it does not by itself exclude the possibility that for certain datasets the calibration can be helpful.
\begin{table}[h!]
	\begin{center}
     
		\resizebox{\linewidth}{!}{%
			\begin{tabular}{@{}llc@{}}
				\toprule
				& Training accuracy & Test accuracy\\ \midrule
				Generative classifier                   & 95.25\%                      & 88.12\%                  \\
				Generative classifier after calibration & 95.43\%                      & 88.15\%                  \\ \bottomrule
		\end{tabular}}
			\caption{Accuracies on the training set and test set before and after calibration.}
	\label{tab:calibration}  
	\end{center}
\end{table}

\begin{table}[h!]
	\centering

	\resizebox{\linewidth}{!}{%
		\begin{tabular}{@{}llllllllll@{}}
			\toprule
			$c_1$ & $c_2$ & $c_3$ & $c_4$ & $c_5$ & $c_6$ & $c_7$ & $c_8$ & $c_9$ & $c_{10}$ \\ \midrule
			-0.068& -0.015&  0.066&  0.067& -0.037&  0.066& -0.068& -0.067& -0.058&  0.062        \\ \bottomrule
		\end{tabular}%
	}
	\caption{The values of the learned   $\{c_1,...,c_K\}$.}
\label{tab:calibrated_vals}
\end{table}

\subsubsection{Model capacity and regularization}
\begin{figure}[h!]
	\centering
	\resizebox{\linewidth}{!}{%
		\centering
		\includegraphics[width=\textwidth]{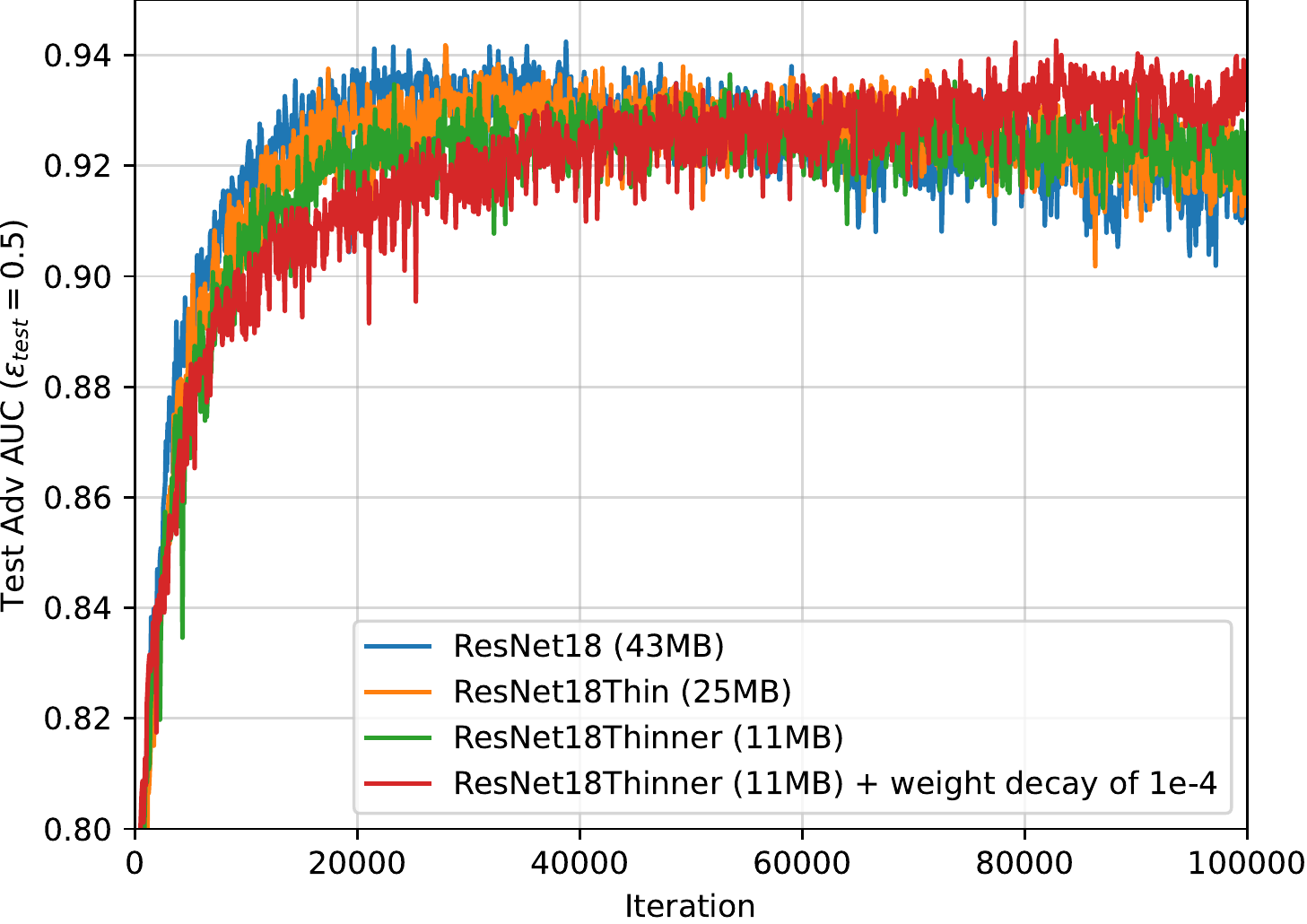}
	}
	\caption{Adversarial AUROC score of the class 0 models of different capacities. ``ResNet18Thin'' and ``ResNet18Thinner'' are two ResNet18 models respectively with a width multiplier of 0.75 and 0.5.} 
	\label{fig:capacity_ablation}
\end{figure}

Consistent with the previous findings~\cite{madry2017towards, alayrac2019labels, xie2019intriguing, gowal2020uncovering}, we find models with higher capacity tend to have better robustness
(\cref{fig:capacity_ablation}). Note that overfitting happened even with ResNet18Thinner, the model with the lowest capacity. To mitigate overfitting, we apply weight decay and \cref{fig:capacity_ablation} shows that the weight decay help the model achieve better robustness.

\subsubsection{Training perturbation size.}
\begin{figure}[h!]
	\centering
	\begin{subfigure}[t]{\linewidth}
		\centering
		\includegraphics[width=\linewidth]{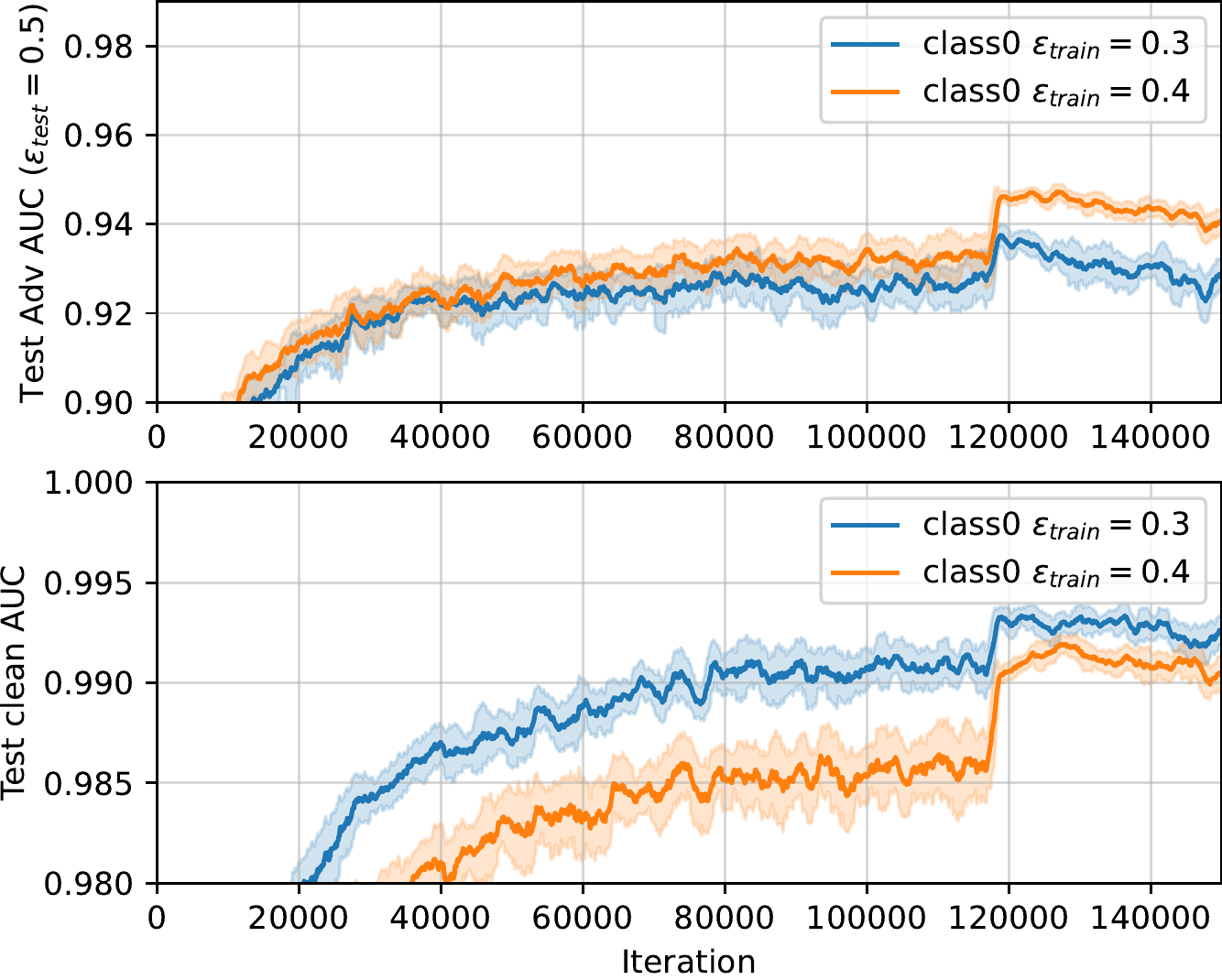}	
	\end{subfigure}
	
	\caption{Standard and adversarial performances of the class 0 model when trained with different perturbation sizes.}
	\label{fig:perturbation_ablation}
\end{figure}

Similar to \cite{tsipras2018robustness}, in \cref{fig:perturbation_ablation} we observe a decline in standard performance as the training perturbation size increases. Meanwhile, training with a larger perturbation helps the model achieve better adversarial robustness.

\subsubsection{Data augmentation}
\begin{figure}[h!]
	\centering
	\resizebox{\linewidth}{!}{%
		\centering
		\includegraphics[width=\textwidth]{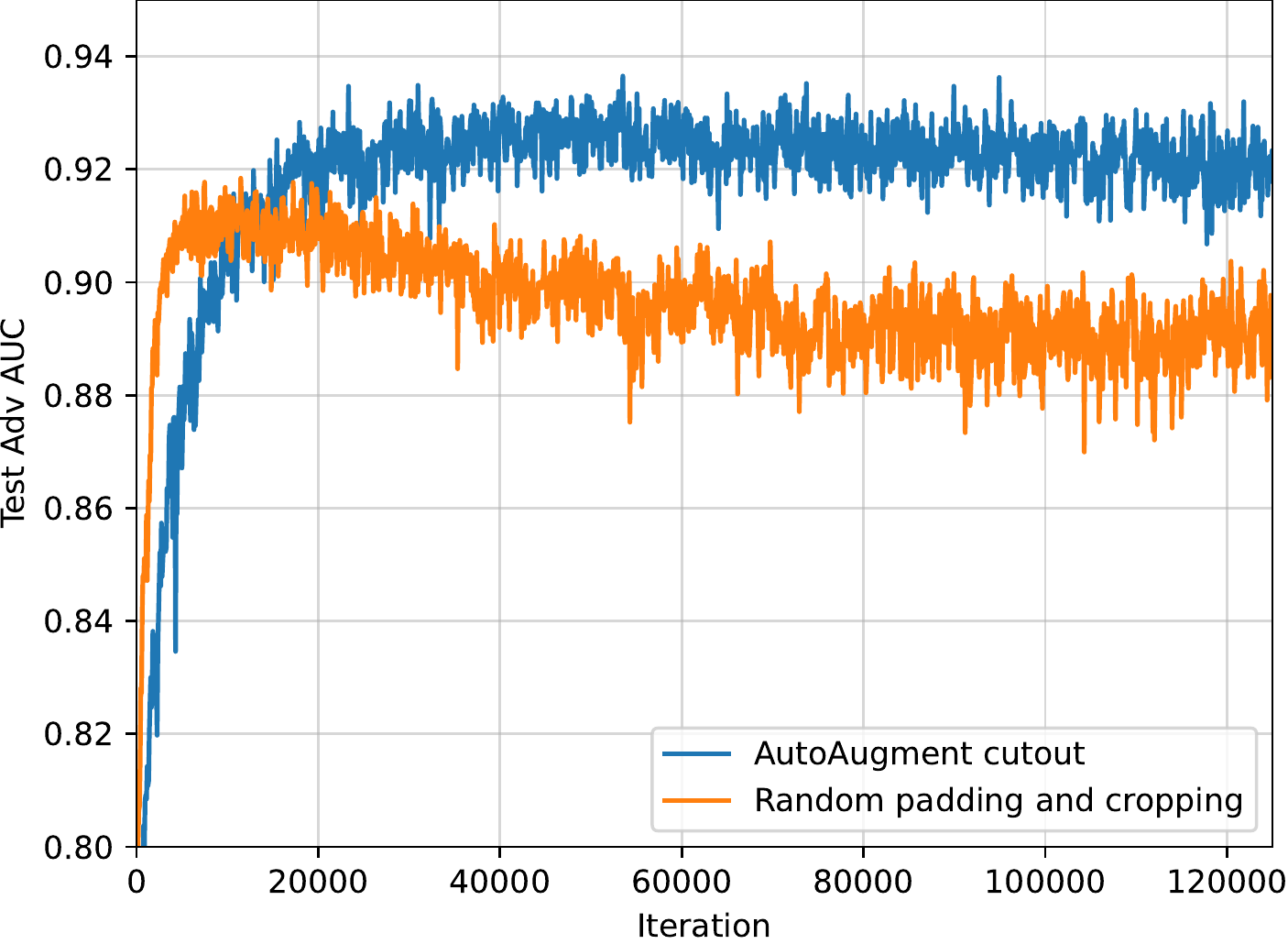}
	}
	\caption{Adversarial AUROC score of the class 0 model when trained with different data augmentation policy.
	}
	\label{fig:aug_ablation}
\end{figure}

Data augmentation is a widely used regularization technique for reducing overfitting in learning standard classification models. Unfortunately, the use of data augmentation in training robust classifiers is not as successful. On CIFAR-10, beyond the widely adopted random padding and cropping augmentation strategy, none of the more sophisticated augmentation techniques are beneficial for improving model robustness~\cite{rice2020overfitting,gowal2020uncovering}.
\cite{rebuffi2021fixing} studies this phenomenon and  find that when combined with model weight averaging~\cite{izmailov2018averaging},  heuristics-driven augmentation techniques such as Cutout~\cite{devries2017improved}, CutMix~\cite{yun2019cutmix} and
MixUp~\cite{zhang2017mixup} can help improve robustness. However, data-driven data augmentation approaches such as AutoAugment~\cite{cubuk2019autoaugment} have not been found to be helpful. In contrast to these work, we find it straightforward to apply AutoAugment in our training to reduce overfitting and obtain better robustness (\cref{fig:aug_ablation}). 

\subsubsection{Combining in- and out-distribution adversarial training}
\begin{figure}[h!]
	\centering
	\resizebox{\linewidth}{!}{%
		\centering
		\includegraphics[width=\textwidth]{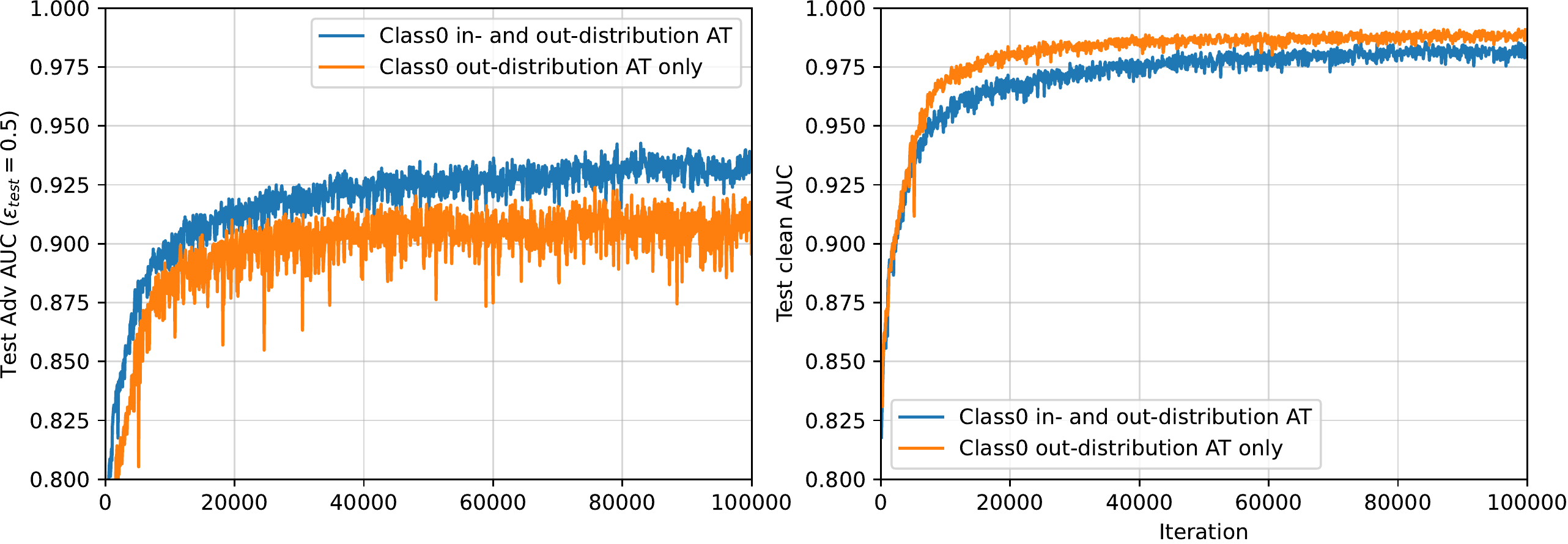}
	}
	\caption{Clean and adversarial AUROC scores of the class 0 model when trained with combined AT and out-distribution AT only. 
	}
	\label{fig:inout_ablation}
\end{figure}

\cref{fig:inout_ablation} shows the effect of combining in- and out-distribution adversarial training. It can be seen that combining in- and out-distribution AT helps the model achieve better adversarial performance at the expense of decreased standard performance.

\subsection{Conclusion}
We have investigated applying the AT generative model to learn class-conditional distributions and then using the conditional models to perform generative classification. Compared to the baseline softmax robust classifier, the generative classifier achieves comparable performance when there is no adversarial perturbation or the perturbation is small, and much better performance when the perturbation is of moderate or large sizes. The generated samples and counterfactuals produced by the generative classifier are also more similar to the training data (as measured by FID) than that of the softmax robust classifier, suggesting that the generative classifier has better captured the class-conditional distributions. We note that the above results are obtained when we use a much  smaller model for the binary classifiers so that the generative classifier and softmax classifier have similar capacities and inference speeds. We further show how model capacity and training perturbation size affect the model performance in a similar way as they would in standard adversarial training. Compared to the softmax robust classifier, we find it much easier to apply advanced augmentation such as AutoAugment in our training. We leave the evaluation of other data augmentation such as CutMix and MixUp to future work. The analysis in \cite{yin2022learning} also suggests that using a diverse out-distribution dataset improve  the generative modeling performance of the binary classifier, but it is unclear whether using auxiliary OOD data helps improve the binary classifier's discriminative capability. Of related work is \cite{augustin2020adversarial} which show that incorporating auxiliary OOD data helps the robust classifier achieves better standard and robust accuracies on CIFAR-10. Further, some recent work show that generated in-distribution data produced by some generative models can be leveraged to achieve better robustness \cite{rebuffi2021fixing,gowal2021improving}.  We leave the investigation of using auxiliary OOD data and/or generated in-distribution data in our approach to future work.

{\small
\bibliographystyle{ieee_fullname}
\bibliography{ref}
}

\end{document}